\documentclass[letterpaper, 10 pt, conference]{ieeeconf}  % Comment this line out if you need a4paper

\IEEEoverridecommandlockouts                              % This command is only needed if 
\usepackage{cite}
\usepackage{graphicx} % for pdf, bitmapped graphics files
\usepackage{amsmath} % assumes amsmath package installed
\usepackage{amssymb}  % assumes amsmath package installed
\usepackage{caption}
\usepackage{arydshln}
\usepackage{algorithm}
\usepackage{algorithmic}
\title{\LARGE \bf
Online Embedding Multi-Scale CLIP Features into 3D Maps
}

\author{Shun Taguchi$^{1}$ and Hideki Deguchi$^{1}$% <-this % stops a space
\thanks{*This work was not supported by any organization}% <-this % stops a space
\thanks{$^{1}$Shun Taguchi and Hideki Deguchi are with Toyota Central R\&D Labs., Inc.,
        44-1 Yokomichi, Nagakute, Aichi, Japan 
        {\tt\small s-taguchi@mosk.tytlabs.co.jp}}%
}

\begin{document}

\maketitle
\thispagestyle{empty}
\pagestyle{empty}

%%%%%%%%%%%%%%%%%%%%%%%%%%%%%%%%%%%%%%%%%%%%%%%%%%%%%%%%%%%%%%%%%%%%%%%%%%%%%%%%
\begin{abstract}
This study introduces a novel approach to online embedding of multi-scale CLIP (Contrastive Language-Image Pre-Training) features into 3D maps. 
By harnessing CLIP, this methodology surpasses the constraints of conventional vocabulary-limited methods and enables the incorporation of semantic information into the resultant maps. 
While recent approaches have explored the embedding of multi-modal features in maps, they often impose significant computational costs, lacking practicality for exploring unfamiliar environments in real time. 
Our approach tackles these challenges by efficiently computing and embedding multi-scale CLIP features, thereby facilitating the exploration of unfamiliar environments through real-time map generation. 
Moreover, the embedding CLIP features into the resultant maps makes offline retrieval via linguistic queries feasible. 
In essence, our approach simultaneously achieves real-time object search and mapping of unfamiliar environments. 
Additionally, we propose a zero-shot object-goal navigation system based on our mapping approach, and we validate its efficacy through object-goal navigation, offline object retrieval, and multi-object-goal navigation in both simulated environments and real robot experiments. 
The findings demonstrate that our method not only exhibits swifter performance than state-of-the-art mapping methods but also surpasses them in terms of the success rate of object-goal navigation tasks.
\end{abstract}

%%%%%%%%%%%%%%%%%%%%%%%%%%%%%%%%%%%%%%%%%%%%%%%%%%%%%%%%%%%%%%%%%%%%%%%%%%%%%%%%
\section{Introduction}

% Background
Efficient navigation in complex environments poses a fundamental challenge for autonomous systems and has been a subject of extensive study \cite{thrun1998probabilistic, endres2012evaluation}. 
Mapping, in particular, has attracted considerable attention due to its critical role in planning and localization. 
In recent years, there has been a growing interest in augmenting maps with semantic information to enhance their usefulness. 
However, traditional approaches relying on object detection \cite{yolov8} and semantic segmentation \cite{chaplot2020object} often operate within fixed vocabularies, limiting their ability to capture the diverse semantics inherent in real-world environments.

% Problem Statement
We contend that maps searchable by open vocabulary queries will enable the performance of a broader range of tasks in complex environments. 
However, existing methods frequently grapple with computational complexity and real-time applicability. 
The computational demands of populating a map with multimodal features present a challenge for real-time search in unknown environments. 
Additionally, techniques that embed single-scale features at specific points constrain the utility of current mapping methodologies, lacking the flexibility to accommodate queries across different scales. 
For instance, a sink in a kitchen may represent a `sink' at a small scale but a `kitchen' at a broader scale. 
Thus, even a single object naturally exhibits different feature values depending on the viewing scale. 
Consequently, there is an urgent need for approaches that not only compute and embed semantic information efficiently but also ensure real-time performance and adaptability to various search contexts at different scales.

\begin{figure}
\includegraphics[width=1.0\hsize]{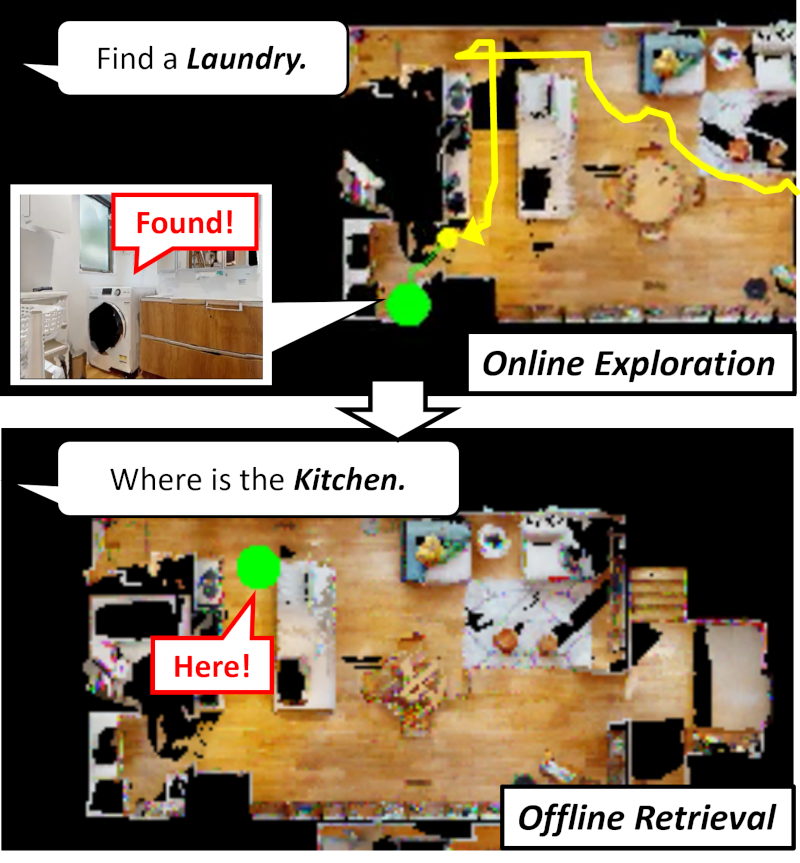}
\caption{{\bf Online Embedding Multi-Scale CLIP Features into 3D Maps.}
Our approach facilitates real-time object exploration in unknown environments through efficient computation.
Additionally, embedding CLIP features into the resulting map enables offline retrieval from the map post-creation, thus augmenting the practical utility of the proposed method.}
\label{fig:pull}
\end{figure}

% Proposition
To tackle this challenge, we present a novel method for efficiently embedding multi-scale CLIP (Contrastive Language-Image Pre-Training) \cite{radford2021learning} features into 3D maps. 
By harnessing the capabilities of visual language models like CLIP, our approach aims to surpass the vocabulary limitations inherent in traditional methods. 
Moreover, we enhance the quality and semantic richness of the map by adopting a multi-scale approach to extracting CLIP features from the images. 
Consequently, our method can retrieve not only objects such as `sink' but also queries referencing spaces like `kitchen'. 
Additionally, by concatenating multi-scale images and efficiently computing and embedding CLIP features in batches, our method achieves real-time object retrieval and mapping in unknown environments. 
Since the resulting map is embedded with features, offline retrieval using linguistic queries becomes feasible, thereby enhancing the practical utility of the map (see Fig. \ref{fig:pull}).

% Contribution
The contributions of this study can be summarized as follows: 
\begin{itemize}
\item We propose a novel and efficient method for real-time embedding of multi-scale CLIP features into 3D maps.
\item We develop a zero-shot object-goal navigation system for mobile robots using our mapping method.
\item We validate the effectiveness of our method through various open-vocabulary tasks, including object-goal navigation, offline object retrieval on maps, and multi-object-goal navigation, in both simulated environments and real robot experiments.
\end{itemize}
We have verified that our method outperforms not only object-goal navigation employing semantic maps based on conventional object detection, such as YOLOv8\cite{yolov8} and Detic\cite{zhou2022detecting}, but also the state-of-the-art offline method, VLMap \cite{huang23vlmaps}.

\section{Related Work}

% Semantic Mapping Techniques
\noindent{\bf Semantic Mapping.}
With the maturation of Visual SLAM (Simultaneous Localization and Mapping) technology and advancements in semantic understanding capabilities driven by deep learning, research on semantic mapping has seen significant growth \cite{salas2013slam++, mccormac2017semanticfusion}. 
Various approaches have been proposed, including object detection-based methods \cite{salas2013slam++, runz2018maskfusion, mccormac2018fusion++, xu2019mid} and semantic segmentation-based techniques \cite{mccormac2017semanticfusion, chaplot2020object}. 
Chaplot et al. \cite{chaplot2020object} introduced object-goal navigation using semantic mapping, demonstrating the feasibility of object exploration in unknown environments by projecting semantic segmentation onto a 2D map. 
However, these semantic mapping techniques, which embed specific categories without vocabulary scalability, often fall short in capturing the full semantic richness of the environment.

% Zero-shot models
\noindent{\bf Open Vocabulary Approaches.}
The emergence of CLIP has transformed the fusion of visual and textual data, offering a powerful means to comprehend the world holistically \cite{radford2021learning}. 
Inspired by the success of CLIP , object detection \cite{zhou2022detecting, kamath2021mdetr} and semantic segmentation \cite{li2022languagedriven} in open vocabularies have been actively explored. 
Detic \cite{zhou2022detecting} is an open vocabulary object detection method utilizing CLIP features and trained on over 21,000 categories of images. 
LSeg \cite{li2022languagedriven} is a semantic segmentation model capable of providing class labels for each pixel as CLIP features. 
Additionally, several methods for object-goal navigation have been proposed using visual language models \cite{khandelwal2022simple, majumdar2022zson, gadre2023cows, yokoyama2024vlfm}. 
CoW \cite{gadre2023cows} employs CLIP to explore for objects from space, whereas 
VLFM \cite{yokoyama2024vlfm} utilizes BLIP-2 \cite{li2023blip} to map the probability of an object's existence, thus enhancing exploration efficiency. 
However, these methods primarily support online exploration, and the resulting maps are often tailored to specific queries. 
The feature embedding method targeted in this study is more versatile as it enables the retrieval of objects from the obtained maps.

% Embedding Multi-Modal Features into Maps
\noindent{\bf Visual-Language Feature Embedded Maps.}
Recent research has directed its focus towards embedding visual-language features into 3D maps to enhance their semantic content. 
While these methods have demonstrated promising results in generating semantically enriched maps, they may encounter challenges related to computational complexity and scalability \cite{huang23vlmaps, lerf2023, Peng2023OpenScene}. 
LM-Nav \cite{shah2022lmnav} offers visual language navigation by embedding CLIP features in the nodes of topological maps and connecting them to landmarks derived from language instructions. 
However, it necessitates pre-mapping, and feature embedding is confined to nodes on the topological map, resulting in sparsity. 
VLMap \cite{huang23vlmaps} employs an open vocabulary semantic segmentation model, LSeg \cite{li2022languagedriven}, to embed features into 2D maps, facilitating navigation for complex queries such as "between the sofa and the TV". 
Nevertheless, the computational overhead of map generation is considerable, and its utilization mandates pre-exploration for map generation. 
LERF \cite{lerf2023} adopts a Neural Radiance Field \cite{mildenhall2021nerf} approach to embed CLIP features into a 3D map. 
Despite producing highly detailed maps, it proves unsuitable for navigation in unknown environments due to the prerequisite pre-computation. 
Some methodologies have been proposed to compute the CLIP features of vertices from the 3D model \cite{Peng2023OpenScene}; however, even these are not tailored for processing point clouds in real-time during exploration. 
While existing methodologies offer precise semantic mapping with an open vocabulary, they frequently encounter high computational overheads, rendering them impractical for real-time or large-scale environment operations. 
In contrast, the method proposed in this study facilitates online map construction while simultaneously enabling retrieval capabilities. 
This approach ensures adaptability to dynamic environmental changes, thereby enhancing its suitability for real-world deployment across diverse scenarios.

\section{Method}
Our proposed method aims to efficiently embed multi-scale CLIP \cite{radford2021learning} features into 3D maps, facilitating the integration of rich semantic information for enhanced map interpretation and object retrieval capabilities. 
The method comprises several key steps: A) Embedding multi-scale CLIP features into 3D maps; B) object retrieval on the generated map; and C) system implementation of object-goal navigation using our mapping method. 
Each step is described in the following subsections, respectively.

\subsection{Embedding Multi-Scale CLIP Features into 3D Maps}
\begin{figure}
\includegraphics[width=1.0\hsize]{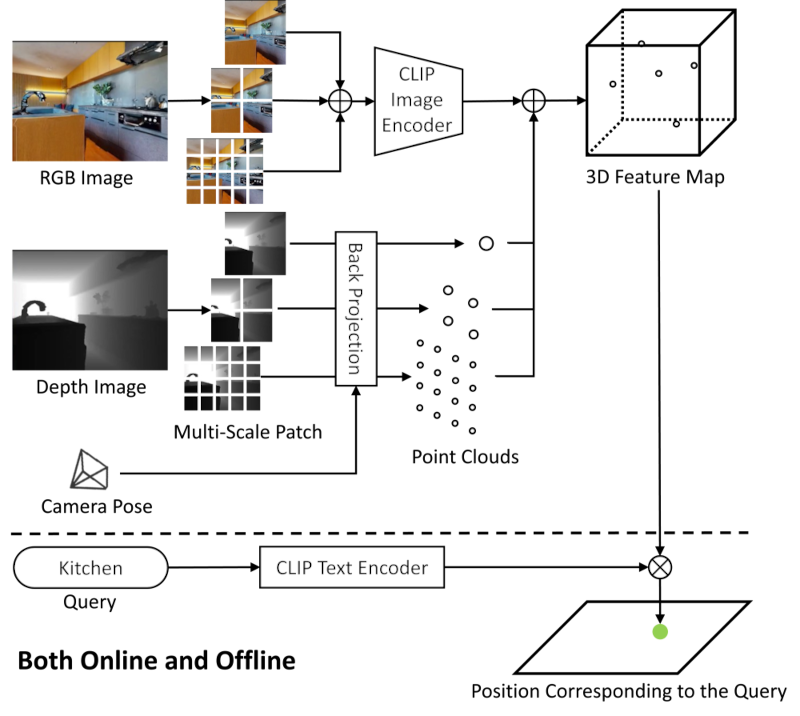}
\caption{{\bf Embedding multi-scale CLIP features into a 3D map.}}
\label{fig:clipmapper}
\end{figure}

The intuitive concept behind the proposed method is to embed CLIP features extracted from RGB images into 3D point locations determined by depth back-projection based on RGB-D camera inputs. 
However, CLIP typically outputs only one feature from each image and does not extract features for partial images. 
To address this limitation, VLMap \cite{huang23vlmaps} employs LSeg \cite{li2022languagedriven}, capable of extracting per-pixel features, for feature embedding. 
However, VLMap encounters high computational demands, confining its application to offline mapping exclusively. 
In contrast, our method divides the image into patches at various scales, concatenating these images in the batch dimension and inputting them to CLIP. 
This approach allows for the acquisition of multi-scale features denser than the entire image yet sparser than pixel-level features. 
Furthermore, by concatenating images in the batch direction and input, the CLIP encoder undergoes computation only once, ensuring efficiency. 
The embedding and retrieval procedure of our proposed method is depicted in Fig. \ref{fig:clipmapper}. 
The procedures are detailed below.

% Feature extraction
Firstly, the observed RGB image $I_t$ at the current time step $t$ is partitioned into patches at multiple scales. 
The size of the patches for each scale $i$ is $s_i \times s_i$, where $s_i = 2^i S$ and $S \times S$ represents the size of the input image of the CLIP image encoder.
The observed image is divided to maximize the number of patches obtainable from its size. 
Specifically, $N_i = \lfloor w/s_i \rfloor \times \lfloor h/s_i \rfloor$, where $w$ and $h$ denote the width and height of the observed image. 
Consequently, the image is center-cropped to a size of $\lfloor w/s_i \rfloor s_i \times \lfloor h/s_i \rfloor s_i$. 
All obtained patch images $I_t(b)$ (where $b$ denotes the index of the patch image) are resized to $I'_t(b)$, with dimensions $S times S$, and concatenated in the batch dimension. 
This enables the CLIP image encoder $\mathcal{F}_{\rm CLIP}^I$ to obtain CLIP features $f_t(b)$ for each resized patch image $I'_t(b)$ concatenated in the batch dimension with only one calculation.
\begin{align}
f_t^I(b) = \mathcal{F}_{\rm CLIP}^I(I'_t(b)).
\end{align}

% Calculating point cloud
To embed the computed CLIP features into 3D points, a point cloud in world coordinates is computed from the depth image and camera pose. 
Initially, to compute the local point cloud, all depth pixels $D_t(p)\in D_t$ are back-projected according to the following formula:
\begin{align}
Q_t(p) = D_t(p) K^{-1} \left[\begin{array}{c}
p \\ 
1
\end{array}
\right],
\end{align}
where $Q_t(p) \in Q_t$ represents the back-projected point, $p = [u, v]^T$ denotes pixel coordinates, and $K$ is the intrinsic matrix of the depth image. 
We assume that the intrinsic matrix of the depth image is identical to that of the RGB image. 
Point clouds in world coordinates are obtained by converting the local point clouds based on the camera pose $P_t \in \mathbb{R}^{4\times4}$:
\begin{align}
\left[\begin{array}{c}
Q_t^{g}(p) \\ 
1 
\end{array}\right] = P_t \left[\begin{array}{c}
Q_{t}(p) \\
1
\end{array}\right],
\end{align}resulting
where $Q_t^{g}$ is a point cloud in world coordinates. 
The obtained point cloud $Q_t^g$ is averaged over the points in each patch image $I_t (b)$ to compute the coordinates embedding the features from each patch image:
\begin{align}
\mathcal{Q}_t^g (b) = \frac{1}{s_i^2} \sum_{p \in I_t(b)}{Q_t^g(p)},
\end{align}
where $\mathcal{Q}_t^g (b)$ is the embedded point for the $b$th patch image $I_t(b)$. 
The calculated points and features are accumulated as set of tuples $\left( \mathcal{Q}_t^g(b), f_t^I(b) \right)$ over time, as follows:
\begin{align}
\mathcal{M}^{\rm CLIP}_{t} &= \mathcal{M}^{\rm CLIP}_{t-1} \cup \left\{ \left( \mathcal{Q}_t^g(b), f_t^I(b) \right) \left| \forall b \right. \right\}, 
\end{align}
where $\mathcal{M}^{\rm CLIP}_{t} = \left( \mathcal{Q}^g, f^I \right)_{0:t}$ represents the accumulated points and features during time steps $0$ to $t$.

\subsection{Object Retrieval on the Generated Map}
\label{sec:retrieval}
Object retrieval is conducted by assessing the similarity between the query and the features embedded in the map. 
Initially, the query $q_t$ is transformed into text features $f_t^q$ using CLIP text encoder $\mathcal{F}_{\rm CLIP}^q$:
\begin{align}
f_t^q = \mathcal{F}_{\rm CLIP}^q (q_t).
\end{align}
Similarity can be computed by computing the cosine similarity between the text features $f_t^q$ and the features $f_{0:t}^I$ embedded in the map:
\begin{align}
\mathcal{S}_{q_t}(e) = \frac{f_t^q \cdot f_{0:t}^I(e)}{\left\|f_t^q\right\| \left\| f_{0:t}^I(e) \right\|},
\end{align}
where $\mathcal{S}_{q_t}(e)$ denotes the similarity between text and point features, and $e$ is an element of the map. 
Based on the obtained similarity, it is determined if the point corresponds to the query using a threshold judgment:
\begin{align}
\mathcal{Q}_{q_t}^g = \left\{ \mathcal{Q}^g(e) \left| \mathcal{S}_{q_t}(e) > \theta \right. \right\},
\end{align}
where, $\mathcal{Q}_{q_t}^g$ denotes the set of points corresponding to the query $q_t$, and $\theta$ represents the threshold of the similarity.

\begin{figure*}
\centering
\includegraphics[width=1.0\hsize]{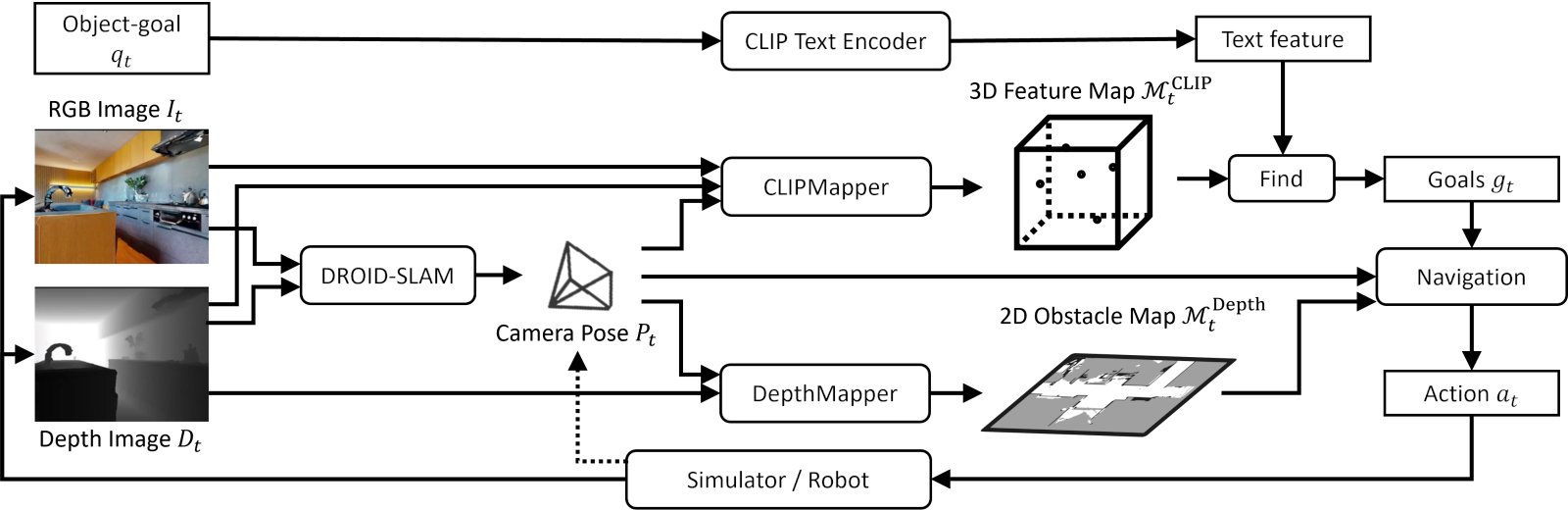}
\caption{{\bf System implementation of object-goal navigation based on the proposed CLIP feature embedding method.}}
\label{fig:implementation}
\end{figure*}

\begin{algorithm}
\caption{Object-goal Navigation}
\label{alg:objecgoal_nav}
\begin{algorithmic}
\STATE \textbf{Input:} Object-goal $q_t$
\STATE \textbf{Observe:} RGB image $I_t$, Depth image $D_t$
\STATE \textbf{Output:} Action $a_t$
\STATE $P_t \leftarrow {\rm DroidSLAM}(P_{t-1}, I_t, D_t)$
\STATE $\mathcal{M}^{\rm Depth}_{t} \leftarrow {\rm DepthMapper}(\mathcal{M}^{\rm Depth}_{t-1}, D_t, P_t)$
\STATE $\mathcal{M}^{\rm CLIP}_{t} \leftarrow {\rm CLIPMapper}(\mathcal{M}^{\rm CLIP}_{t-1}, I_t, D_t, P_t)$
\STATE $g_t \leftarrow {\rm Find}(\mathcal{M}^{\rm CLIP}_{t-1}, q_t, \theta)$
\IF{$g_t \neq \emptyset$}
    \STATE $a_t \leftarrow {\rm Navigation}(\mathcal{M}^{\rm Depth}_{t}, g_t, P_t)$
\ENDIF
\IF{$g_t = \emptyset$ \OR $a_t = \emptyset$}
    \STATE $a_t \leftarrow {\rm Navigation}(\mathcal{M}^{\rm Depth}_{t}, g_{\rm distant}, P_t)$
    \WHILE{$a_t = \emptyset$}
        \STATE $\theta \leftarrow \theta - 0.001$
        \STATE $g_t \leftarrow {\rm Find}(\mathcal{M}^{\rm CLIP}_{t-1}, q_t, \theta)$
        \IF{$g_t \neq \emptyset$}
            \STATE $a_t \leftarrow {\rm Navigation}(\mathcal{M}^{\rm Depth}_{t}, g_t, P_t)$
        \ENDIF
    \ENDWHILE
\ENDIF
\RETURN $a_t$
\end{algorithmic}
\end{algorithm}

\subsection{System Implementation of Object-Goal Navigation Using Our Mapping Method}
We establish a zero-shot open-vocabulary object-goal navigation system using our mapping method. 
The data flow is illustrated in Fig. \ref{fig:implementation}, and the algorithm is presented in Algorithm \ref{alg:objecgoal_nav}. 
In this system, the query $q_t$, indicating the object-goal, serves as input, whereas the RGB image $I_t$ and depth image $D_t$ are observed from the RGB-D camera to execute navigation to the target object. 
Initially, the camera pose $P_t$ is estimated from the RGB-D image using DROID-SLAM \cite{teed2021droid}, utilizing the pre-trained DROID-SLAM model without additional training. 
Subsequently, a 3D feature map $\mathcal{M}^{\rm CLIP}_t$ is generated by the proposed mapping system (CLIPMapper in Fig. \ref{fig:implementation}), and a 2D obstacle map $\mathcal{M}^{\rm Depth}_t$ is derived from the depth image $D_t$ and camera pose $P_t$ by projecting a point cloud onto a 2D map (DepthMapper in Fig. \ref{fig:implementation}). 
Goals $g_t$ corresponding to the query $q_t$ are retrieved from the 3D feature map $\mathcal{M}^{\rm CLIP}_t$ by the procedure in Sec.\ref{sec:retrieval} (Find in Algorithm \ref{alg:objecgoal_nav}). 
The navigation module (navigation in Fig. \ref{fig:implementation}) searches for the shortest path to a goal $g_t$ by utilizing multi-goal A* on the obstacle map $\mathcal{M}^{\rm Depth}_t$ and outputs an action $a_t$ to follow the path. 
If the goal $g_t$ is empty or unreachable, navigation proceeds by using a specific point $g_{\rm distant}$ as the goal to explore the environment. 
In the event that exploration fails to find a path to a distant point $g_{\rm distant}$ due to complete space closure, a gradually decreasing similarity threshold $\theta$ is employed to search for reachable goals.

\section{Experiments}

\subsection{Experimental Setup}
This study verifies the effectiveness of the proposed method through simulation and real robot experiments. We employ the Habitat \cite{habitat19iccv, szot2021habitat} simulator for simulation and the Vizbot \cite{niwa2022spatio} platform for real robot experiments.

\noindent{\bf Habitat} \cite{habitat19iccv,szot2021habitat} is a simulation platform tailored for training embodied AI agents in photo-realistic 3D environments. 
It offers high-fidelity simulation capabilities, allowing agents to navigate and interact in diverse virtual spaces. 
Our evaluation of the proposed method in Habitat utilizes a model house environment captured by the Matterport Pro2 camera \cite{matterport}. 
The position and rotation of the 2D robot can be observed in the simulations.

\noindent{\bf Vizbot} \cite{niwa2022spatio}, developed at Toyota Central R\&D Labs., Inc., is a compact mobile robot platform derived from the Roomba. 
It is equipped with a Richo Theta S 360° camera, a RealSense D435i RGB-D camera, and Hokuyo's Lidar. 
The robot incorporates an Intel NUC as a computing device and communicates with a computation server via Wi-Fi using ROS. 
For our study, we solely utilize RGB-D cameras as sensors, with robot poses estimated by DROID-SLAM \cite{teed2021droid} from RGB-D observations.

\noindent{\bf Common Settings.}
The input RGB-D image sizes are configured as $640 \times 480$ pixels, whereas CLIP models process images of size $224 \times 224$ pixels. 
Scale parameters for calculating multi-scale CLIP features are set as $i = [1, 0, -1]$. 
The size of cells in the 2D map is designated as $5$ cm. 
The distant point $g_{\rm distant}$ is positioned at $(-50, -50)$, outside the floors. 
For both experiments, primary computations were conducted on a desktop PC equipped with an Intel Xeon 3.60GHz CPU and NVIDIA TITAN RTX GPU. 
In the robot experiments, Wi-Fi-based remote control from the desktop PC was utilized.

\subsection{Baselines}
To assess the performance of our method, we conducted comparisons with several detection and mapping techniques, employing the same navigation algorithm for all methods.

\noindent{\bf YOLOv8 \cite{yolov8}} is a recent object detection method, which provides many pre-trained models that include semantic segmentation.
In this study, we utilized a pre-trained segmentation model (yolov8x-seg.pt) to generate a semantic map by projecting it onto a 2D map based on depth information for object localization. 
The model supports 80 categories of objects trained by COCO dataset \cite{lin2014microsoft} and can detect objects within those categories with high accuracy.
The confidence threshold parameter for object detection was set at $0.7$.

\noindent{\bf Detic \cite{zhou2022detecting}} is an object detection approach trained on over 21,000 categories of data. 
Leveraging CLIP features for category embeddings, it exhibits a degree of generalizability to language. 
Detic also offers semantic segmentation capabilities. 
We employed a pre-trained model to construct a semantic map by projecting it onto a 2D map, similar to the approach for YOLOv8. 
The confidence threshold parameter for object detection was set at $0.7$.

\noindent{\bf VLMap \cite{huang23vlmaps}} stands as the state-of-the-art semantic mapping method utilizing LSeg to embed vision-language features into maps. 
VLMap operates offline, requiring prior exploration of the environment via our exploration method to generate maps from collected data. 
Navigation is then executed using the map. 
However, for evaluating the map's performance, we utilized the same navigation algorithm as employed in our methods.

\subsection{Object-goal Navigation}

The experiments were conducted using Habitat \cite{habitat19iccv, szot2021habitat} as the experimental platform. 
The experimental environment consists of two distinct model room environments consists of two floors, and a total of 16 types of objects (bed, chair, sofa, toilet, dining table, tv, sink, refrigerator, bathtub, laundry, kitchen, stairs, hallway, living room, carpet, and window) are searched for.
Among these, the first 8 objects are included in COCO \cite{lin2014microsoft}, are detectable by YOLOv8 \cite{yolov8}, whereas the others are not included in COCO.
Objects not included in COCO encompass not only specific objects like `bathtubs' and `laundry' but also spaces such as `kitchen' and `living room'. 
Each episode of the experiment was dedicated to one of the existing objects across four different environments, amounting to a total of 54 episodes. 
An episode is deemed successful if, upon stopping, the robot is within 2.0m of any target object and can see it by turning.
If no stop action is selected within 1000 steps, the episode is deemed a failure.

\begin{table}
\begin{center}
\caption{{\bf Success ratio (SR) of object-goal navigation with various detection method.}
Our method has an extremely high success rate for objects not included in COCO \cite{lin2014microsoft} compared to other methods, indicating that it has acquired a high generalizability to object variations.}
\label{tab:objnav_results}
\begin{tabular}{ll|c|c|c|c}
\hline
   & & & & \multicolumn{2}{c}{SR for each object group} \\ 
\multicolumn{2}{l|}{Method} & online & SR & Obj. in COCO \cite{lin2014microsoft} & Others \\
\hline
\multicolumn{2}{l|}{YOLOv8\cite{yolov8}} & \checkmark & 25.9 & 53.8 & 0.0 \\
\multicolumn{2}{l|}{Detic\cite{zhou2022detecting}} & \checkmark & 29.6 & 50.0 & 10.7 \\
\multicolumn{2}{l|}{VLMap\cite{huang23vlmaps}}  & & 50.0 & 46.2 & 53.6 \\
\hdashline[1pt/1pt]
{\bf Ours} & Vit-B/32 & \checkmark & 64.8 & 61.5 & 67.9 \\ 
 & Vit-L/14 & \checkmark & {\bf 87.0} & {\bf 80.8} & {\bf 92.9} \\
\hline
\end{tabular}
\end{center}
\end{table}

\begin{figure}[th]
\centering
\includegraphics[width=1.0\hsize]{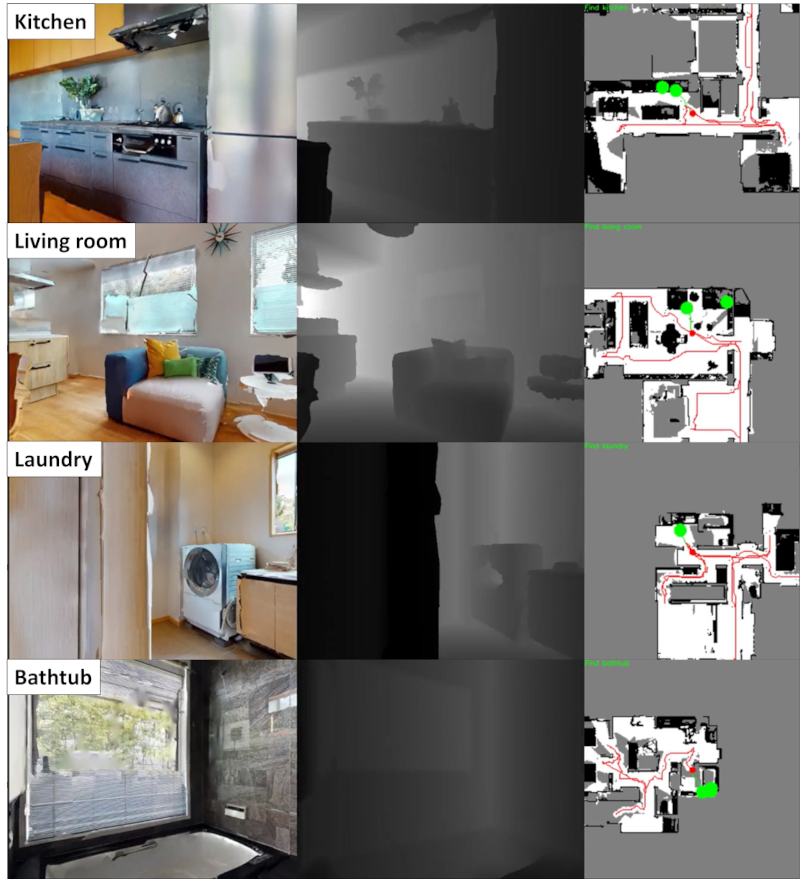}
\caption{
{\bf Qualitative results of object-goal navigation using our mapping method (ViT-L/14).} 
From left to right: RGB image, depth image, and 2D obstacle map.
The red circle on the map indicates the robot's position, the red line represents the robot's trajectory, the green circle denotes the detected object's position, and the green line marks the path to the object's position.
This demonstrates our method's ability to locate objects specified by arbitrary text.}
\label{fig:objnav_results}
\end{figure}

The results of the Object-goal Navigation evaluation by our method and baselines are presented in Table \ref{tab:objnav_results}. 
Additionally, the results of our method with different CLIP models (ViT-B/32, ViT-L/14) are compared. 
For the ViT-B/32 model, the initial similarity threshold was set to 0.3, whereas for ViT-L/14, it was set to 0.27. 
The findings reveal that the success rate (SR) of our method surpasses those of the baselines, YOLOv8 \cite{yolov8}, Detic \cite{zhou2022detecting}, and VLMap \cite{huang23vlmaps}. 
Although YOLOv8 exhibits a moderate SR for objects in COCO \cite{lin2014microsoft}, it fails to detect objects beyond its vocabulary. 
While Detic offers open-vocabulary object detection, in this context, only bathtubs were detectable among objects not included in COCO, with no other objects being detectable. 
VLMap demonstrates high navigation accuracy in an open vocabulary setup owing to semantic mapping by LSeg. 
However, even the ViT-B/32 model of our method outperforms VLMap. 
This discrepancy is likely attributed to false positives, as VLMap navigates to the nearest objects with detailed segmentation on a pixel-by-pixel basis. 
In contrast, our approach has a higher probability of reaching the correct object because it finds objects with sufficient similarity immediately and navigates toward the point of maximum likelihood when no object with sufficient similarity.

Fig. \ref{fig:objnav_results} illustrates the actual observations and the generated maps when the proposed method detected objects that were not included in COCO \cite{lin2014microsoft}.
The figure demonstrates the proposed method's capability to detect objects specified by arbitrary text in the environment.

Additionally, we assess the effectiveness of multi-scale CLIP features through an ablation study. 
Table \ref{tab:multi-scale_ablation} presents the results of the ablation study regarding the scales of CLIP features. 
This table displays the SR and features per image at each scale. 
It is evident that patches that are too small exhibit poor detection performance. 
This finding confirms that a higher SR can be achieved by employing multi-scale CLIP features rather than single-scale features.

\begin{table}
\begin{center}
\caption{{\bf Ablation study on the effectiveness of embedding multi-scale CLIP features.}
We use ViT-L/14 of CLIP models.}
\label{tab:multi-scale_ablation}
\begin{tabular}{l|cccc|c|c}
\hline
             & \multicolumn{4}{c|}{Scales} & & \\
Method       & 1 & 0 & -1 & -2             & feat./image & SR \\
\hline
Single-Scale & \checkmark & & & & 1 & 75.9 \\
             & & \checkmark & & & 4  & 77.8 \\
             & & & \checkmark & & 20 & 72.2 \\
             & & & & \checkmark & 88 & 51.9 \\
\hdashline[1pt/1pt]
{\bf Ours (Multi-Scale)} & \checkmark & \checkmark & & & 5 & 79.6 \\
 & \checkmark & \checkmark & \checkmark & & 25 & {\bf 87.0}\\
 & \checkmark & \checkmark & \checkmark & \checkmark & 113 & 85.2 \\
\hline
\end{tabular}
\end{center}
\end{table}

The computational time of each mapping method is detailed in Table \ref{tab:computational_time}. 
Here, we have implemented VLMap \cite{huang23vlmaps} as an online method and measured its computational time. 
As indicated in the table, our method is 60 times faster than VLMap. 
The total computation time of our method using the ViT-L/14 model is approximately 100 msec, comparable to mapping with YOLOv8 \cite{yolov8} or Detic \cite{zhou2022detecting}, which is a realistic computation time for performing online mapping and exploration simultaneously. 
Furthermore, our model using the ViT-B/32 model requires only 20 msec for mapping and retrieval, indicating its potential for applications that demand more real-time performance.

\begin{table}
\begin{center}
\caption{{\bf Computational time [sec] required by each mapping method.}
VLMap\cite{huang23vlmaps} has been implemented as the online method and its computational time has been measured.}
\label{tab:computational_time}
\begin{tabular}{ll|c|c|c}
\hline
\multicolumn{2}{l|}{Method} & Total & Mapping & Retrieval \\
\hline
\multicolumn{2}{l|}{YOLOv8\cite{yolov8}} & 0.13 & 0.06 & 0.07 \\
\multicolumn{2}{l|}{Detic\cite{zhou2022detecting}} & 0.13 & 0.12 & {\bf 0.01} \\
\multicolumn{2}{l|}{VLMap (online)\cite{huang23vlmaps}} & 7.00 & 1.20 & 5.80 \\
\hdashline[1pt/1pt]
 % & RN50 &  &  &  \\
{\bf Ours} & Vit-B/32 & {\bf 0.02} & {\bf 0.01} & {\bf 0.01} \\
           & Vit-L/14 & 0.12 & 0.02 & 0.10 \\
\hline
\end{tabular}
\end{center}
\end{table}

\subsection{Offline Object Retrieval}
The evaluation proceeds with object retrieval. 
In this experiment, a closed space exploration is conducted using a object-goal navigation described earlier without specifying a particular object-goal.
By calculating the similarity between arbitrary text and the 3D CLIP feature map generated by this system and visualizing it, offline object retrieval on the map is executed.

\begin{table}
\begin{center}
\caption{{\bf Precision@1 on offline object retrieval.}}
\label{tab:retrieval_results}
\begin{tabular}{ll|c|c|c}
\hline
                          & & & \multicolumn{2}{c}{P@1 for each object group} \\ 
\multicolumn{2}{l|}{Method} & P@1 & Obj. in COCO \cite{lin2014microsoft} & Others \\
\hline
VLMap\cite{huang23vlmaps} & Raw  & 50.0 & 46.2 & 53.6 \\
 & Normalized & {\bf 85.2} & {\bf 88.5} & 
82.1 \\
\hdashline[1pt/1pt]
{\bf Ours} & Vit-B/32 & 74.1 & 69.2 & 78.6 \\
 & Vit-L/14 & 83.3 & 80.8 & {\bf 85.7} \\
\hline
\end{tabular}
\end{center}
\end{table}

\begin{figure}
\includegraphics[width=1.0\hsize]{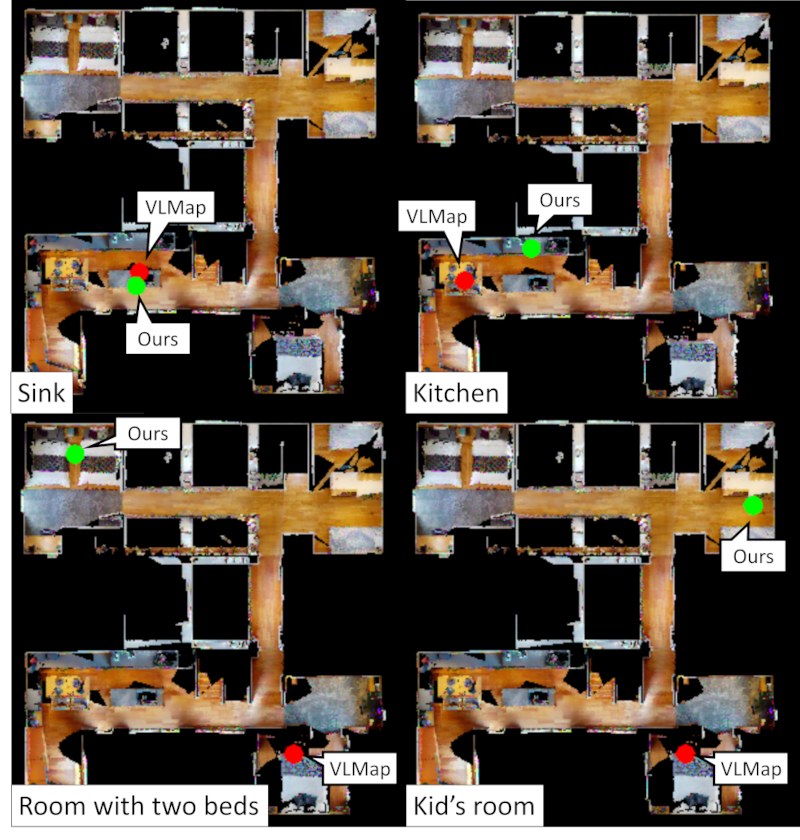}
\caption{{\bf Comparison results of object retrieval.}
The results suggest that our method exhibits better retrieval ability for spatial queries than VLMap.
}
\label{fig:space_queries}
\end{figure}

Object retrieval is assessed by the SR (precision@1) of the highest scoring point location for each object. 
A result is considered successful if it is situated within 50 cm of an instance of the target object. 
Although VLMap does not offer a method for retrieving the highest scoring positions, two methods were evaluated here: one that outputs the highest LSeg score itself (Raw) and one that uses scores normalized by the `other' class scores (Normalized). 
Table \ref{tab:retrieval_results} presents the precision@1 of object retrieval. 
While VLMap (Normalized) demonstrates high performance, it is evident that the map obtained by our method achieves comparable performance in an efficient manner. 
Additionally, VLMap exhibits greater accuracy for individual objects, such as those included in COCO\cite{lin2014microsoft}, whereas our method excels for more abstract queries.

Fig. \ref{fig:space_queries} illustrates the comparison results of offline object retrieval between VLMap(Normalized) and our method. 
While both VLMap and our method successfully detect object queries such as `sink', our method surpasses VLMap in detection accuracy for queries indicating space, such as `kitchen'.
Moreover, when searching for more complex spatial queries such as `room with two beds' or `kid's room', our method accurately identifies the correct location, whereas VLMap struggles. 
These results affirm that our method enables offline object retrieval on generated maps and indicate that it outperforms VLMap in spatial queries.

\subsection{Multi-Object-Goal Navigation}

\begin{table}
\begin{center}
\caption{{\bf Results on multi-object-goal navigation.}
The numbers represent the number of episodes succeeded in reaching each subgoal.}
\label{tab:multion_results}
\begin{tabular}{ll|c|cccc}
\hline
   & & & \multicolumn{4}{c}{No. of Subgoals in Row} \\ 
\multicolumn{2}{l|}{Method} & online & 1 & 2 & 3 & 4 \\
\hline
\multicolumn{2}{l|}{VLMap\cite{huang23vlmaps}} & & 52 & 31 & 18 & 14 \\
\hdashline[1pt/1pt]
%{\bf Ours} & ViT-B/32 & \checkmark & 67 & 36 & 23 & 13 \\ % scale [-2, -1, 0, 1] 
{\bf Ours} & ViT-B/32 & \checkmark & 68 & 35 & 20 & 12 \\ % scale [-1, 0, 1]
%& ViT-L/14 & \checkmark & 95 & 65 & 45 & 38 \\            % scale [-2, -1, 0, 1] 
& ViT-L/14 & \checkmark & {\bf 96} & {\bf 65} & {\bf 44} & {\bf 37} \\            % scale [-1, 0, 1] 
\hline
\end{tabular}
\end{center}
\end{table}

We also assessed multi-object-goal navigation, a task more complex than object-goal navigation, combining aspects of both object-goal navigation and offline object retrieval. In multi-object-goal navigation within an unknown space, the optimal strategy involves effectively exploring unknown areas while utilizing previously mapped regions. In this study, multi-object-goal navigation was executed for four randomly chosen object-goals, and the number of achieved subgoals was measured. The assessment encompassed four environments, each comprising 25 episodes, totaling 100 episodes.

The results are displayed in Table \ref{tab:multion_results}. Here, we compared our method with the leading competitor in object-goal navigation, VLMap \cite{huang23vlmaps}. 
Our method (ViT-L/14) succeeded in reaching the fourth object in more than twice as many episodes as VLMap.
The results show that our method outperforms VLMap even in multi-object-goal navigation.

\begin{figure}
\includegraphics[width=1.0\hsize]{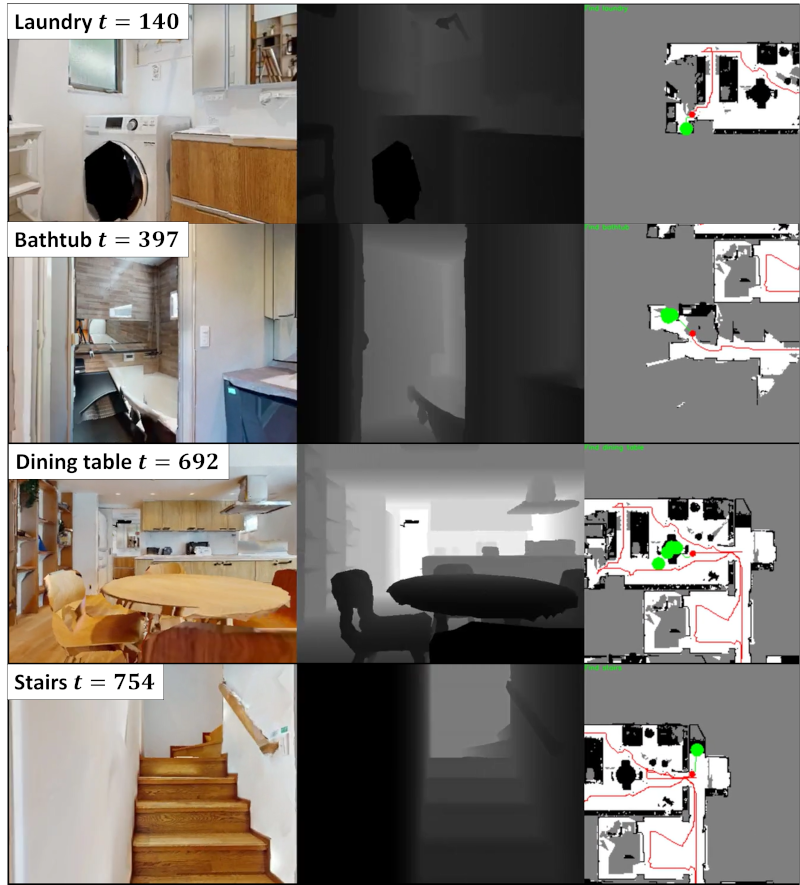}
\caption{{\bf Qualitative Result of Multi-object-goal Navigation using our mapping method (ViT-L/14).}
Since our method can embed CLIP features online in the map, we can confirm that online exploration and retrieval on the map can be performed simultaneously.
}
\label{fig:multion}
\end{figure}

The Fig. \ref{fig:multion} presents qualitative results from a task where the user continuously searches for a chair, kitchen, hallway, and sofa in a simulation environment.
Our method embeds CLIP features into the map, enabling concurrent exploration for map construction and object retrieval based on previous observations. When given a query for an object absent from the map, the system explores unknown regions. 
Conversely, if the object exists on the map (e.g., dining table and stairs in Fig. \ref{fig:multion}), navigation is directed toward the identified point on the map. 
These results affirm that our method adeptly integrates online exploration and retrieval on the map to achieve multi-object-goal navigation successfully.

\subsection{Real Robot Experiments}

\begin{figure}
\includegraphics[width=1.0\hsize]{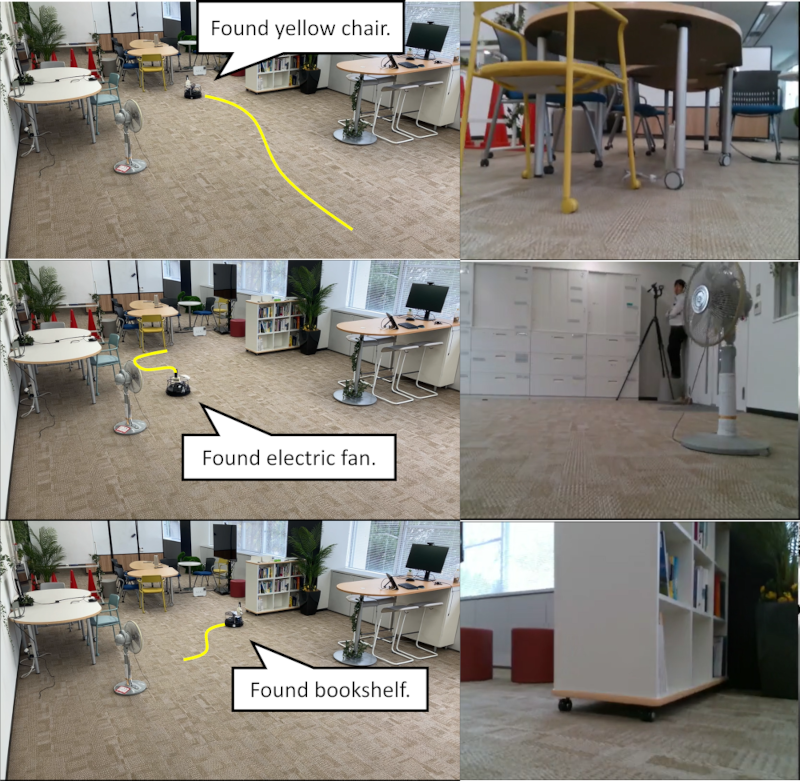}
\caption{{\bf Multi-object-goal Navigation Results on Real Robot.}
The left image is from an external camera and the right image is from a camera mounted on the robot.
The yellow line indicates the trajectories of the robot.
}
\label{fig:real_multion}
\end{figure}

Multi-object-goal navigation was also conducted using an actual mobile robot, Vizbot \cite{niwa2022spatio}. 
DROID-SLAM \cite{teed2021droid} was employed for the robot's localization. 
Fig. \ref{fig:real_multion} illustrates the outcomes of multi-object-goal navigation on Vizbot using our method. 
In this experiment, we deployed our method (ViT-L/14) in an office environment and tasked the robot with locating a yellow chair, electric fan, and bookshelf. 
The results demonstrate our method's ability to navigate to multiple objects based on language queries in a real-world scenario. 
This observation confirms that our method can concurrently generate a map with embedded CLIP features while exploring, even when implemented on a real robot, enabling navigation to objects instructed via language commands. 
Given that our method operates on a zero-shot architecture, it circumvents the sim2real problem and can seamlessly adapt to real robots without requiring additional training.

\section{Conclusion}
This study introduces a novel approach for efficiently embedding multi-scale CLIP features into 3D maps, addressing the limitations of traditional vocabulary-constrained methods. 
By harnessing the capabilities of CLIP, our approach facilitates the integration of diverse semantic information into the generated maps, thereby enhancing their interpretability and utility. 
Through comprehensive experimental validation, including object-goal navigation, offline object retrieval, and multi-object-goal navigation, we have demonstrated the effectiveness and versatility of our proposed methodology.

One limitation of our method is the sparsity of the provided maps. 
While this is acceptable for tasks not requiring centimeter-level accuracy of goals, such as object-goal navigation, it may be insufficient for tasks like picking. 
Another limitation is that, since our system focuses on navigation to the maximum likelihood point in space, it may overlook the nearest object, resulting in longer travel paths. 
Thus, there is a trade-off with false positives, underscoring the importance of setting appropriate parameters.

A potential future direction is the integration with LLM. 
Employing LLM to extract landmarks \cite{deguchi2024language} or generating code \cite{liang2023code} from language instructions could enable complex tasks such as visual language navigation.
Additionally, exploring techniques to enhance the scalability and efficiency of our approach in large-scale environments is crucial for practical deployment.

In summary, our research represents a significant advancement in the development of more efficient and semantically rich 3D mapping techniques, with potential applications in various fields such as robotics, autonomous navigation, and geographic information systems.

\section*{ACKNOWLEDGMENT}
We extend our gratitude to N. Hirose, K. Sukigara, and K. Ito from Toyota Central R\&D Labs., Inc. for their invaluable cooperation in collecting the data of 3D the model house environment used in this study.

\bibliographystyle{IEEEtran}
\bibliography{ref}

\end{document}